\documentclass[
]{ceurart}

\usepackage{listings}
\usepackage{latexsym}
\usepackage{amssymb}
\usepackage{amsmath}
\usepackage{amsthm}
\usepackage{booktabs}
\usepackage{enumitem}
\usepackage{graphicx}
\usepackage{color}
\usepackage{makecell}
\usepackage{xcolor,colortbl}
\usepackage{array,multirow}
\usepackage{comment}
\usepackage{url}
\usepackage{caption}
\usepackage{subcaption}
\usepackage{siunitx}
\usepackage{rotating}
\usepackage{epigraph}
\usepackage{arydshln}

\begin{document}

\copyrightyear{2025}
\copyrightclause{Copyright for this paper by its authors.
  Use permitted under Creative Commons License Attribution 4.0
  International (CC BY 4.0).}

\conference{Identity-Aware AI workshop at 28th European Conference on Artificial Intelligence,
  October 25, 2025, Bologna, Italy}

\title{Who are you, ChatGPT? Personality and Demographic Style in LLM-Generated Content}


\author[]{Dana Sotto Porat}[%
email=dana.sotto@gmail.com,
]
\cormark[1]
\address[]{The Academic College of Tel Aviv--Yaffo, Tel Aviv--Yaffo, Israel}

\author[]{Ella Rabinovich}[%
email=ellara@mta.ac.il,
]


\begin{abstract}
Generative large language models (LLMs) have become central to everyday life, producing human-like text across diverse domains. A growing body of research investigates whether these models also exhibit personality- and demographic-like characteristics in their language. In this work, we introduce a novel, data-driven methodology for assessing LLM personality without relying on self-report questionnaires, applying instead automatic personality and gender classifiers to model replies on open-ended questions collected from Reddit. Comparing six widely used models to human-authored responses, we find that LLMs systematically express higher Agreeableness and lower Neuroticism, reflecting cooperative and stable conversational tendencies. Gendered language patterns in model text broadly resemble those of human writers, though with reduced variation, echoing prior findings on automated agents. 
We contribute a new dataset of human and model responses, along with large-scale comparative analyses, shedding new light on the topic of personality and demographic patterns of generative AI.

\end{abstract}


\maketitle


\section{Introduction}
\label{sec:introduction}

People differ in their personality, and these differences have been shown to be expressed in language \citep{sun2018personality, peters2024large, shemtov2025exploring}. Subtle cues in word choice, tone, and style can reveal aspects of one's underlying traits, making language a valuable window into character and personality. Generative AI is increasingly shaping both personal and professional experiences, capable of managing knowledgeable discussions while also simulating human-like conversational style.

Among the most widely used frameworks for assessing personality are the Big Five traits: \textbf{O}penness (OPN), \textbf{C}onscientiousness (CON), \textbf{E}xtroversion (EXT), \textbf{A}greeableness (AGR), and \textbf{N}euroticism (NEU), collectively abbreviated as "OCEAN". Originally introduced by \citet{goldberg1992development}, this framework has guided extensive research in psychology. More than a decade of computational studies has further shown that personality is reflected in linguistic production (to the extent detectable by automatic tools), motivating the development of techniques for personality assessment from language \citep{greenberg2015musical, sun2018personality, peters2024large}.

In this study, we ask whether generative LLMs --- models trained on vast and diverse corpora --- produce language that spans a range of personality and demographic characteristics resembling those of humans, when used in their most "natural" setting.\footnote{Where the models are \textbf{not} prompted to simulate any particular personality, but rather reply in their most "natural" way.} Previous studies have approached this question by adapting human self-report questionnaires to LLMs: models are asked personality inventory items (e.g., "You often feel easily annoyed or irritable.") and respond on a 5-point accurate--inaccurate scale. Their responses are then scored with the same mappings applied to humans \citep{serapio2023personality, jiang2023evaluating, sorokovikova2024llms, salecha2024large, hartley2025personality, bhandari2025evaluating}. However, this self-report methodology has been criticized \citep{gupta2023self, dorner2023personality} for presupposing that LLMs possess a stable inner nature, rather than merely generating plausible answers (see Section~\ref{sec:related-work} for details).

We instead attempted at an unbiased approach, automatically detecting LLMs’ personality traits along the OCEAN dimensions from their generated language. Specifically, we collected a set of open-ended questions from topical Reddit\footnote{\url{https://www.reddit.com/}} threads -- questions that naturally elicit descriptive, expressive answers. We then gathered responses from both Reddit users and multiple LLMs prompted to reply as if they were social media authors. These responses were analyzed using automatic tools for personality and gender detection, enabling controlled comparison between human and model outputs.

Demographic traits such as gender have also been shown to manifest in language, to the extent detectable by automatic classifiers (see \citet{hacohen2022survey} for comprehensive survey). We therefore extend our analysis to examine whether LLMs’ responses reflect gender likelihood distributions similar to those of human authors.

Our results, based on three open-source and three closed-source models, show that LLMs systematically exhibit higher Agreeableness and lower Neuroticism, likely reflecting their cooperative and psychologically stable training objectives. We also found that gendered language in model outputs broadly aligns with human patterns, though with slightly reduced variation, echoing findings on limited demographic diversity of social spambots \citep{giorgi2021characterizing}.

The contributions of this work are twofold. First, we collect and release a curated dataset of open-ended questions together with both human and model responses, designed to elicit rich, expressive language. Second, we apply a novel large-scale approach for extracting personality traits of generative LLMs along the Big Five dimensions, offering new insights into the personality- and demographic-like qualities of AI text. All our data and code are available at \url{https://github.com/danasotto/llm-personality}.

\section{Related Work}
\label{sec:related-work}

\subsection{Automatic Personality Detection from Language}
The study of personality has historically been the domain of psychology, where researchers have proposed a variety of theories to capture and explain stable behavioral traits in humans. Among these, the Big Five framework \citep{de2000big} and Cattell’s Sixteen Personality Factors (16PF) model \citep{cattell2001sixteen} stand out as particularly influential. Both have been shown to offer consistent and reliable descriptions of individual differences and have therefore been widely adopted in empirical studies. Indeed, decades of research have demonstrated that personality traits correlate with a wide range of real-world behaviors \citep{raad2002big}, and that such traits are also reflected in people's everyday language use \citep{norman1963toward, mehl2006personality}.

\paragraph{Personality of Generative LLMs}
In recent years, a growing body of research has studied the question whether generative LLMs can also be said to exhibit "personality", typically operationalized in terms of the Big Five OCEAN inventory. The prominent methodology involves adapting human self-report questionnaires: models are presented with personality inventory items (questions), and their responses are then scored using the same mappings applied to humans \citep{serapio2023personality, jiang2023evaluating, sorokovikova2024llms, salecha2024large, hartley2025personality, bhandari2025evaluating}. Consider example question, assessing the EXT trait, from the Machine Personality Inventory (MPI, \citep{jiang2023evaluating}), in which models are prompted as follows (similarly to humans):

\begin{quote}
Given the statement: "You feel comfortable around people." please choose the option that best describes you. Options:

(A) Very Accurate \\
(B) Moderately Accurate \\
(C) Neither Accurate Nor Inaccurate \\
(D) Moderately Inaccurate \\
(E) Very Inaccurate
\end{quote}

Responses are then mapped onto trait scores, e.g., selecting (A) would indicate a high level of Extroversion. Aggregating responses across many such items allows researchers to infer an LLM’s personality profile, in a way analogous to human self-report studies. Findings suggest that LLMs tend to score relatively high on Agreeableness and Conscientiousness, with more variable outcomes for the traits of Openness, Extroversion, and Neuroticism.

Further work has shown that LLMs are not fixed in their profiles: they can be induced, through carefully crafted prompts, to adopt different personality configurations, such as a more extroverted or more neurotic persona \citep{jiang2023evaluating, sorokovikova2024llms, zhu2024personality}. This flexibility raises questions about whether such evaluations are measuring anything intrinsic to the model, or merely reflecting surface-level adaptations to instructions. Indeed, the use of self-report questionnaires for models has been criticized on precisely these grounds \citep{gupta2023self}. Unlike humans, LLMs do not possess stable inner states, so "answering" such questions may be more about simulating a plausible response than revealing an underlying disposition. \citet{dorner2023personality} highlights this critique, arguing that "measurement models that are valid for humans do not fit for LLMs, and that currently applied procedures for administering questionnaires to LLMs do not allow for the inference of personality."

Our work proposes an alternative approach: rather than relying on self-reported questionnaires, we assess LLM personality through their more "spontaneous" linguistic productions. Echoing methods long established in psycholinguistic research, we analyze how models respond to a carefully collected set of real-world questions, capture traces of personality that "shine through" in natural language use, and compare them to those found in humans.

\subsection{Automatic Gender Detection from Language}
Differences in language use between men and women have long been a focus of sociolinguistics and gender studies. Robin Lakoff's foundational work "Language and Woman's Place" \citep{lakoff1973language} argued that language reflects, and reinforces, broader gendered social and cultural structures. Subsequent work has expanded and nuanced this claim, documenting the ways in which male (M) and female (F) speakers may differ in their linguistic choices across contexts \citep{labov1990intersection, coates1998language}. Computational research has since provided large-scale empirical confirmation of these trends: across domains and genres, men and women's language often differs systematically, to the point that relatively simple classifiers can achieve robust accuracy in predicting gender from text (for a comprehensive survey, see \citet{hacohen2022survey}).

\paragraph{Demographics of Generative LLMs}
In contrast to the well-developed literature on gender detection in human-authored language, there has been relatively little research on probing the gendered characteristics of generative LLMs. A handful of studies suggest that LLMs exhibit a tendency toward male-coded language \citep{kotek2023gender, soundararajan2024investigating}, a result that is perhaps unsurprising given that a considerable ratio of training corpora are produced by men. These findings highlight how demographic imbalances in training data can manifest in the stylistic and pragmatic profiles of generated text.

Most closely related study was conducted by \citet{giorgi2021characterizing}, who examined social spambots -- automated models producing text for social media platforms, and compared their linguistic characteristics to those of genuine human users. They found, among others, that spambots expressed limited variation along demographic axes such as gender and age, and displayed narrower emotional repertoires. At the same time, spambots tended to overproduce positive sentiment compared to humans. While these models are not as advanced as today's LLMs, the study underscores the ways in which generated text can diverge systematically from human baselines.

Building on this insight, our work advances the literature by conducting a large-scale, controlled evaluation of contemporary LLMs, both open- and closed-source. We seek to provide a more rigorous account of the implicit gender-linked "signature" that emerge in LLM-generated language, and to assess the extent to which these signature resembles patterns observed in human populations.

\section{Datasets Collection}
\label{sec:dataset}

We study the question of LLMs' personality through a comparative analysis of traits extracted from texts authored by human writers and those found in generative model replies. Specifically, we first collect a large dataset of open-ended questions (posts) from diverse topical communities on Reddit, along with expressive answers to those questions by human users (comments). Reddit is a large-scale, user-driven online platform that hosts discussions, content sharing, and community interactions across a wide range of topics. Its structure is organized into subreddits --- thematic communities dedicated to specific subjects, interests, or activities --- each governed by its own rules and moderated by community members. Subreddits can range from broad themes such as politics, technology, or health, to highly specialized interests and niche communities.\footnote{Over 22M subreddits were indexed by the Pushshift API\footnote{\url{https://pushshift.io/signup}} in early 2025: \url{https://tinyurl.com/59hp698u}.}

Using a subset of the collected posts, we next query multiple open- and closed-source LLMs, asking them to provide replies to these posts as if they were social media users. This tightly constrained and controlled setting enables a reliable comparative analysis of the traits displayed by models versus those exhibited by humans.
Details on the data collection process are provided below.

\subsection{Collecting Questions and Comments by Redditors}
\label{sec:collecting-posts}
To focus on open-ended questions that invite descriptive answers, we sampled posts from subreddits across diverse domains such as technology, science, health, lifestyle, entertainment, and social issues. Focusing on conversational content, we filtered in posts by predefined flairs --- metadata property indicating a post's nature --- such as \texttt{Question}, \texttt{Ask}, \texttt{Advise}, \texttt{Discussion}, and \texttt{Poll}. We used the freely available Python PRAW (Python Reddit API Wrapper) package,\footnote{\url{https://praw.readthedocs.io/en/stable/}} which provides structured access to Reddit’s API. 
Below are a few examples of collected questions (post titles and their content), taken verbatim from the dataset:

\begin{quote}
"Opinions on Working and Homeschooling: I have seen a lot of individual opinions that you cannot work a full-time jobs and homeschool. [...]" 
\end{quote}

\begin{quote}
"Space Viruses and Microbial Life: If we discover microbial life on another planet, how do you think that would impact society? Would it change your perspective on life?"
\end{quote}

\begin{quote}
"Bodybuilding while still in school? I have a problem. I started cutting and trying to lose weight/bodyfat in the beginning of my summer break and have been able to control pretty much everything I eat, but now school is starting again and where I go to school you aren't allowed to bring own food because we have a school kitchen that cooks for us. [...]"
\end{quote}

Aiming at comments of sufficient length for meaningful personality and demographics analysis, we filtered out those shorter than 100 words or longer than 300 words. Our final dataset comprises 13K posts and over 30K comments, drawn from 175 diverse subreddit communities, authored by thousands of Reddit users. No sociodemographic information about the authors (such as geographical origin or gender) is available through the platform -- Reddit users are publishing anonymously.

\subsection{Generating Comments with AI Models}
Using the collected posts and comments, we solicited responses from LLMs. A subset of posts was used for this purpose, targeting approximately 10K comments in total from each LLM -- a size large enough for robust analysis, while remaining affordable for closed models.

We employed three commercial models, namely GPT4.1 \citep{openai2025gpt4.1}, GPT4.1-mini \citep{openai2025gpt4.1}, and Claude-Sonnet4.0 \citep{anthropic2025claudesonnet4}, as well as three SOTA\footnote{At the time of conducting the experiments.} open models: Llama3.3-70B \citep{meta2025llama3.3-70b}, Mixtral8x22B \citep{mistral2024mixtral8x22b}, and Qwen2.5-72B \citep{alibaba2024qwen2.5-72b}, for our personality experiments. Each model was run under two settings: with the default temperature of zero ($t{=}0.0$) and with an increased temperature of 0.7 ($t{=}0.7$), to assess whether the less restrictive setting would yield more "diverse" personalities. 
All models were prompted with the following concise instructions, designed to minimize bias in their responses. Here, X denotes the number of comments collected from Reddit for the given post; both the title and content of the post were provided:

\begin{quote}
"Behave like several social media users. Generate exactly <X> comments, at least 100 and at most 300 words each, in response to the following post. The comments should differ from each other and be diverse, like if written by different people.

Post title: <the title of the post>

Post body:  <the content>"
\end{quote}


Compliance with the prompt varied across models, with closed models generally more accurate. Some replies required formatting adjustments, and models occasionally missed the requested number of comments, causing totals to exceed or fall slightly short of 10K, though still adequate for analysis. Table 1 reports the final dataset statistics. For human-authored comments, only a portion of the data --- over 11K out of the total 30K --- was used in experiments; we release the full Reddit dataset, in addition to the data summarized in Table 1, to support future research in this field.

Among open models, Mixtral8x22B often fell short of the minimum word count, so we lowered the threshold to 50 words. No clear biases emerged from this adjustment during analysis.

\begin{table}[h!]
\label{tbl:datasets}
\caption{Dataset statistics: total comments collected, minimum word count in a comment and mean comment length.}
\centering
\begin{tabular}{lrrrr}
\toprule
model & temp & total & min(WC) & avg(WC) \\
\midrule
Claude-Sonnet4.0 & 0.0 & 9,940  & 100 & 165.36 \\
Claude-Sonnet4.0 & 0.7 & 8,813  & 100 & 165.96 \\ \hdashline
GPT4.1 & 0.0 & 8,346  & 100 & 127.49 \\
GPT4.1 & 0.7 & 15,505 & 100 & 129.70 \\ \hdashline
GPT4.1-mini & 0.0 & 7,183  & 100 & 125.28 \\
GPT4.1-mini & 0.7 & 7,426  & 100 & 126.29 \\
\midrule
Llama3.3-70B & 0.0 & 16,350 & 100 & 163.21 \\
Llama3.3-70B & 0.7 & 16,186 & 100 & 160.30 \\ \hdashline
Mixtral8x22B & 0.0 & 8,271  & 50  & 75.90 \\
Mixtral8x22B & 0.7 & 8,519  & 50  & 75.27 \\ \hdashline
Qwen2.5-72B & 0.0 & 8,015 & 100 & 138.23 \\
Qwen2.5-72B & 0.7 & 8,904 & 100 & 139.44 \\
\midrule
human authors & --- & 11,678 & 100 & 155.57 \\
\bottomrule
\end{tabular}
\end{table}

\section{Generative AI: Personality Traits}
\label{sec:experiments-big-five}

Automatic personality classification from text is inherently challenging because personality is a complex, multi-dimensional construct that does not map directly onto linguistic cues in a simple or consistent way. Individual differences in writing style, topic choice, and contextual influences such as social setting or medium of communication make it difficult to isolate stable personality markers. Cultural and language-specific variation further complicates the task, as expressions of the same trait may differ widely across populations. Nevertheless, more than a decade of research in this area has produced models of varying complexity and success. Advances in natural language processing and machine learning have enabled the analysis of large-scale datasets, leading to gradual improvements in predictive accuracy, though the task remains challenging.

Extraction of the Big Five personality traits from text is typically cast as a classification problem, where several classifiers have been proposed over the years with differing levels of accuracy, largely due to the scarcity of high-quality training data \citep{shemtov2025exploring}. In this study, we adopt the recently introduced Big Five personality classifier \citep{shemtov2025exploring}, due to its clear benefits for social media text, particularly, training data collected from Reddit discussions. The model consists of a multilingual encoder connected to a logistic regression classifier, which is trained to estimate the likelihood that a given text exhibits a high level of a given trait. For example, a paragraph assigned a score of 0.85 for EXT is interpreted as strongly indicative of Extroversion.

Following the approach in \citet{shemtov2025exploring}, we applied the \texttt{intfloat/e5-large-v2} encoder \citep{wang2022text} to Reddit comments (both human-authored and LLM-generated). We then used the pre-trained classification models for prediction, assigning each text five continuous scores across the five personality trait dimensions.
Table~\ref{tbl:user-text-traits-examples} provides illustrative comments from our dataset (human-written and generated), together with their automatically assigned low and high NEU scores. Recall that Neuroticism is typically associated with negative emotions, emotional instability, low tolerance for frustration, and increased vulnerability to mental health difficulties. The automatically assigned NEU scores for the example comments in Table~\ref{tbl:user-text-traits-examples} are therefore consistent with intuition.

\begin{table}[h!]
\caption{Example (verbatim) posts and their corresponding comments authored by human Redditors, as well as by closed and open LLMs for the NEU trait. The probability score indicates the likelihood of high trait presence (posterior), as produced by the classification model.
}
\centering
\resizebox{\textwidth}{!}{
\begin{tabular}{lp{14cm}r}
\toprule
Reddit post & Memory loss. Anyone else in here at can’t remember shit? I can’t remember short term or long term. Sometimes things will come to me days after. Sometimes it’s literally years before a random memory pops up. Not all trauma memories just simple stuff too. [...] I’m not depressed I just got chronic ptsd and my memory is funky. & \\ \toprule
Reddit author & For sure. My focus and short term memory are not great. Part of it is that I have a very busy life, I have 2 young kids etc etc. I have been recently diagnosed with ADHD and am primarily inattentive type (more classically thought of as ADD), and treating that has made a difference in my focus which I find helps me remember better, but I completely understand where you are coming from. In terms of the antidepressant that your psychiatrist recommended [...] & 0.677 \\ \midrule
Claude-Sonnet4.0 & Holy shit, yes! I thought I was losing my mind until I started researching the connection between PTSD and memory issues. It's actually super common - trauma literally rewires how our brains store and retrieve memories. I have the same thing where I'll randomly remember something from like 5 years ago while I can't remember what I had for breakfast. It's not just the "bad" memories either - it's like my brain just decided to file everything in the wrong cabinet. [...] & 0.714 \\ \midrule
Llama3.3-70B & I feel you, I've been experiencing similar issues with my memory, especially since my car accident a few years ago, and it's like my brain just can't seem to recall simple things, like where I put my keys or what I had for breakfast, and it gets really frustrating, especially when people around me don't understand what I'm going through, they just think I'm being careless or forgetful, but it's not that, it's like my brain is just wired differently now [...] & 0.650 \\ \midrule \midrule
Reddit post & How is everyone doing with the upcoming holidays regarding simple living? We have managed to convince our families to only buy gifts for kids under 18 which will save us nearly \$1,000. Our budget for gifts is around \$150 now! Yay! [...] Happy Holidays everyone! Prioritize your peace this upcoming season. & \\ \toprule
Reddit author & I'm sewing my own Christmas decorations and make wreaths from things I find in my garden. Also, I discovered that baking sweets is a joy. I'm not good at cakes, but I enjoy making cookies. I made a traditional confect with almond paste, candied orange peel, quark, and flour today together with some meringue drops. [...] & 0.376 \\ \midrule
GPT4.1 & This post really resonated with me! My family also made the decision a few years ago to stop buying gifts for adults, and it’s honestly made the holidays so much more enjoyable and stress-free. We used to spend way too much money and time trying to find the "perfect" gifts for everyone, and it always felt a bit forced. Now, like you, we just buy for the kids and focus on spending quality time together. [...] Happy holidays and cheers to slow, peaceful living! & 0.364 \\ \midrule
Qwen2.5-72B & Love the approach you're taking! It's so refreshing to see a family prioritizing simplicity and financial responsibility. We've been trying to do something similar, but it's been a bit of a struggle to get everyone on board. It's great to hear that you've managed to convince your families to focus on gifts for the kids under 18. It's a smart way to keep things meaningful without overspending. [...] Wishing you a peaceful and cozy holiday season! & 0.296 \\
\bottomrule
\end{tabular}
}
\label{tbl:user-text-traits-examples}
\end{table}

\subsection{Evaluation of Personality Detection Results}
We further validate the automatically assigned personality scores by identifying five subreddits with the highest mean score and five with the lowest mean score for each trait. This computation was performed separately for human- and model-generated comments for EXT, OPN, NEU, and AGR. We deliberately exclude the CON trait from this analysis, as Conscientiousness is particularly difficult to infer from text: it often reflects internal attributes such as self-discipline, organization, and reliability, which do not consistently manifest in explicit surface-level word choices. Also, subreddits with fewer than 50 comments were excluded from the analysis.

\begin{figure}[h!]
\centering
\includegraphics[width=\linewidth]{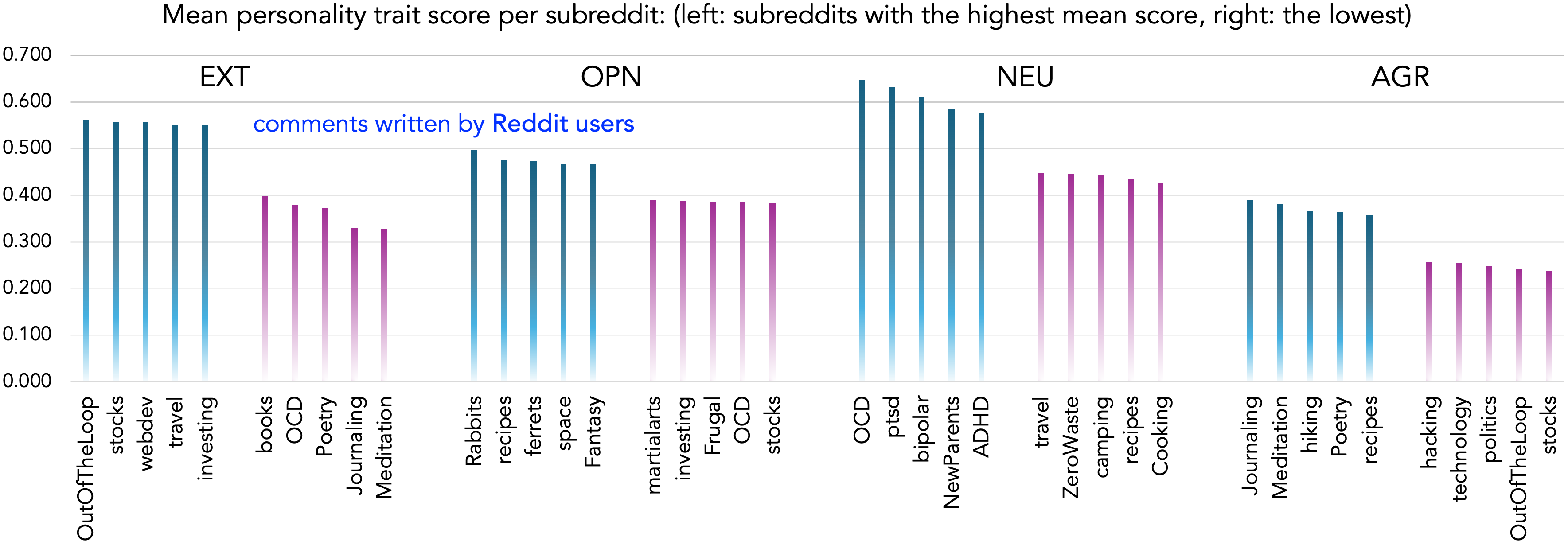}
\caption{Subreddits exhibiting the highest and lowest mean score per trait in comments produced by human Reddit users.}
\label{fig:subreddits-traits-human}
\end{figure}

\begin{figure}[h!]
\centering
\includegraphics[width=\linewidth]{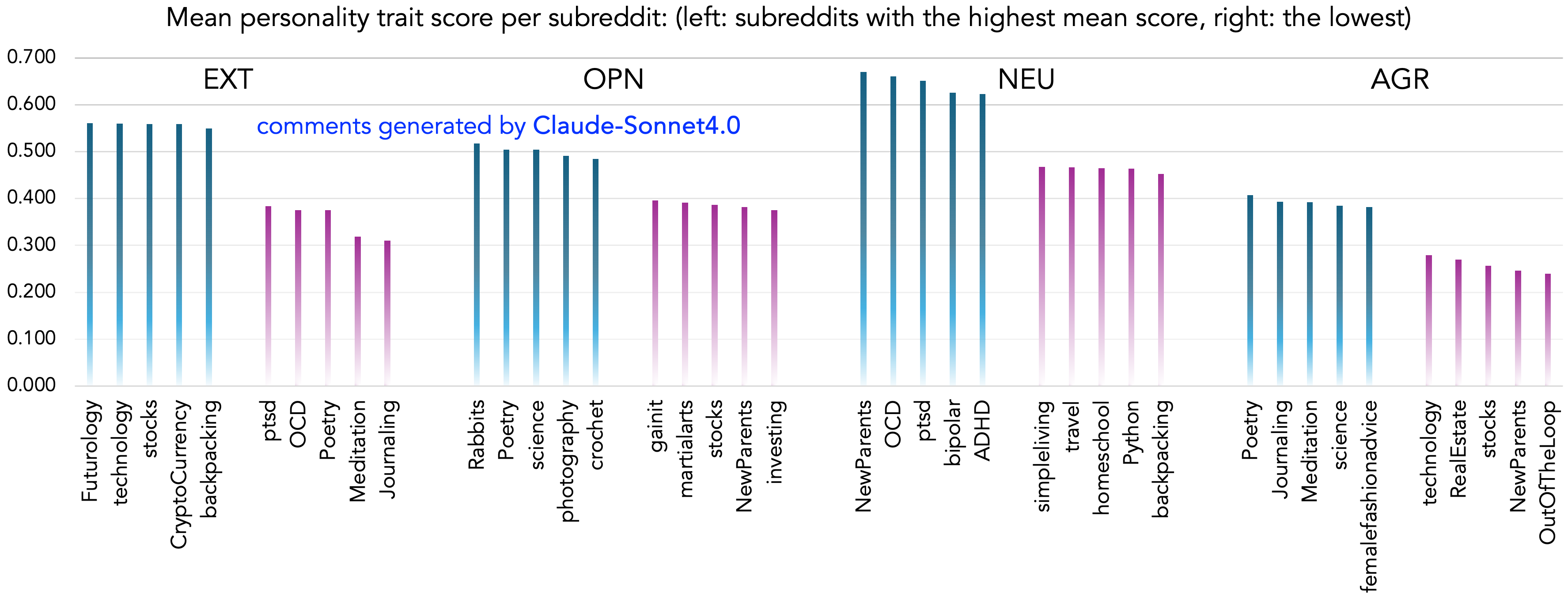}
\caption{Subreddits exhibiting the highest and lowest mean score per trait in comments produced by the Claude-Sonnet4.0 model.}
\label{fig:subreddits-traits-model1}
\end{figure}

\begin{figure}[h!]
\centering
\includegraphics[width=\linewidth]{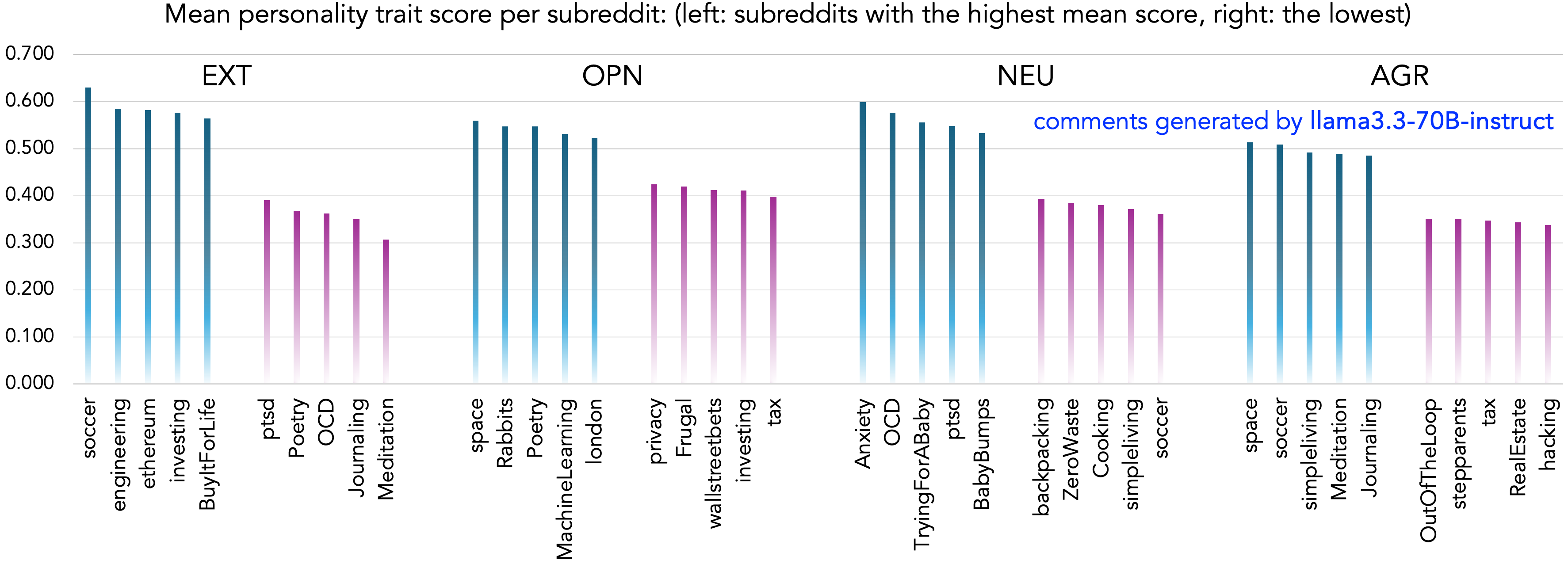}
\caption{Subreddits exhibiting the highest and lowest mean score per trait in comments produced by the Llama3.3-70B model.}
\label{fig:subreddits-traits-model2}
\end{figure}

Figures~\ref{fig:subreddits-traits-human}, \ref{fig:subreddits-traits-model1} and \ref{fig:subreddits-traits-model2} present the results for human- and LLM-generated comments. Careful inspection reveals findings that largely align with intuition. Among Redditors, low EXT comments are concentrated in topical threads such as \texttt{books}, \texttt{OCD}, \texttt{poetry}, \texttt{journaling}, and \texttt{meditation}. Comments with high mean NEU scores appear in \texttt{OCD}, \texttt{ptsd}, \texttt{bipolar}, \texttt{newparents}, and \texttt{ADHD} discussions. The model results also display plausible patterns, with \texttt{simpleliving}, \texttt{homeschool}, and \texttt{backpacking} notable for low NEU in Claude-Sonnet4.0, and \texttt{privacy}, \texttt{frugal}, and \texttt{tax} for low OPN in Llama3.3-70B.

These results suggest that the personality classifier reliably captures the Big Five traits in our data. In the next step, we conduct a comparative analysis of the mean trait levels and their variance across human- and model-written comments.

\subsection{Big Five: Human Authors vs Generative Models}
We compute the mean comment score for each of the Big-Five traits in texts written by human authors and those generated by models. Table~\ref{tbl:results-personality} reports the mean and standard deviation (STD) results. Several insights emerge from these numbers: EXT and OPN mean scores of models are generally comparable to those of human authors, with OPN scores slightly higher. All models exhibit considerably higher AGR scores and lower NEU scores (especially evident in the open models), consistent with prior findings from studies using self-reported questionnaires (see Section~\ref{sec:related-work}), and aligning with the intuition that models are trained to be cooperative, psychologically "stable", and agreeable. Indeed, quite a few of our solicited model responses open with phrases such as "Hey, I totally get where you're coming from!", "I'm so glad you shared this [...]", or "I'm so sorry to hear that you're feeling this way". We do report CON scores in Table~\ref{tbl:results-personality} as well, but refrain from interpreting them. 

\begin{figure}[h!]
\centering
\begin{minipage}[t]{0.48\textwidth}
\includegraphics[width=1\linewidth]{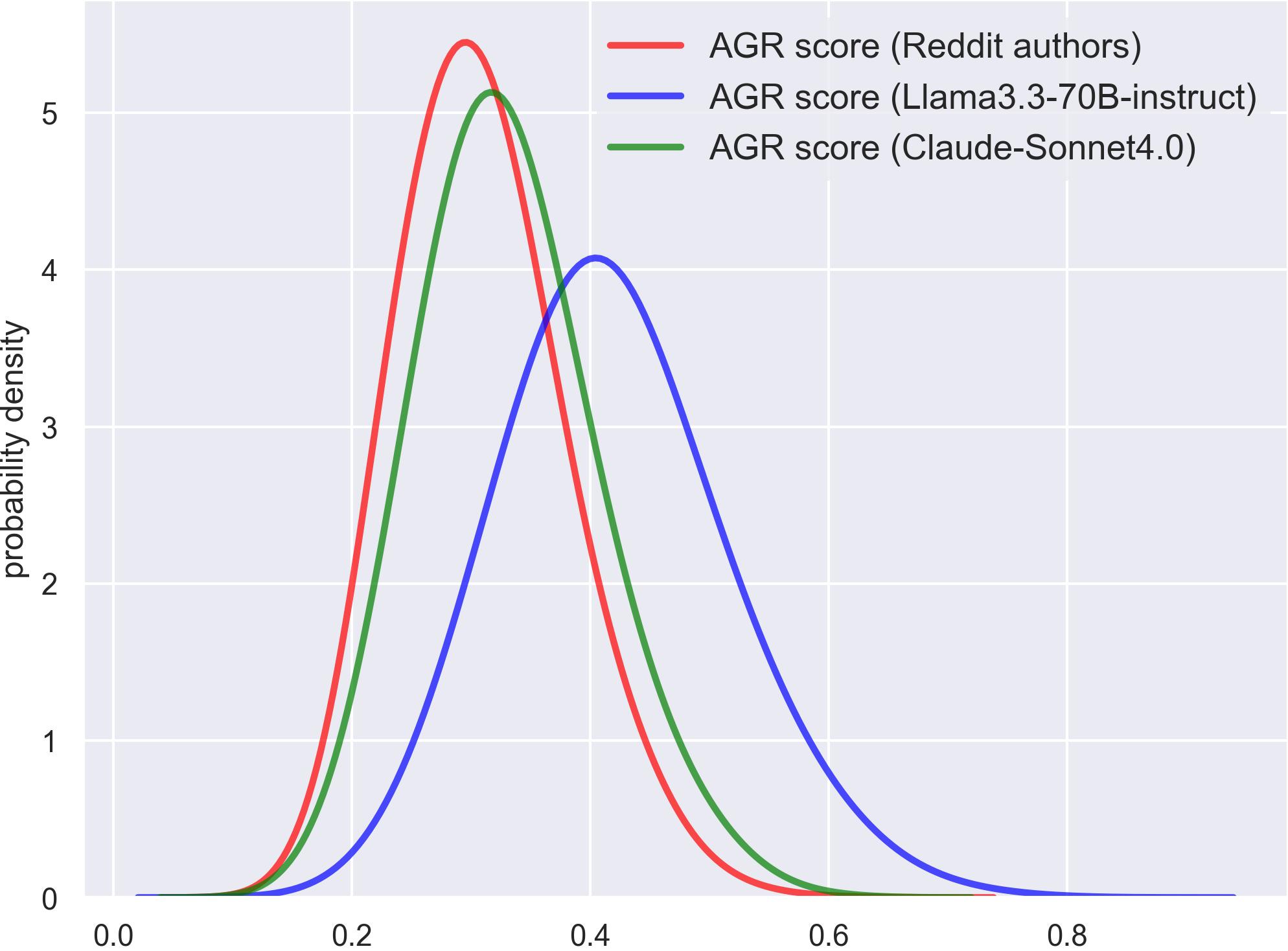}
\caption{Example kernel density distribution \\ of the AGR score in human comments vs LLMs: \\ differences are evident in both mean and STD.}
\label{fig:kernel-agr}
\end{minipage}
\hfill 
\begin{minipage}[t]{0.48\textwidth}
\centering
\includegraphics[width=1\linewidth]{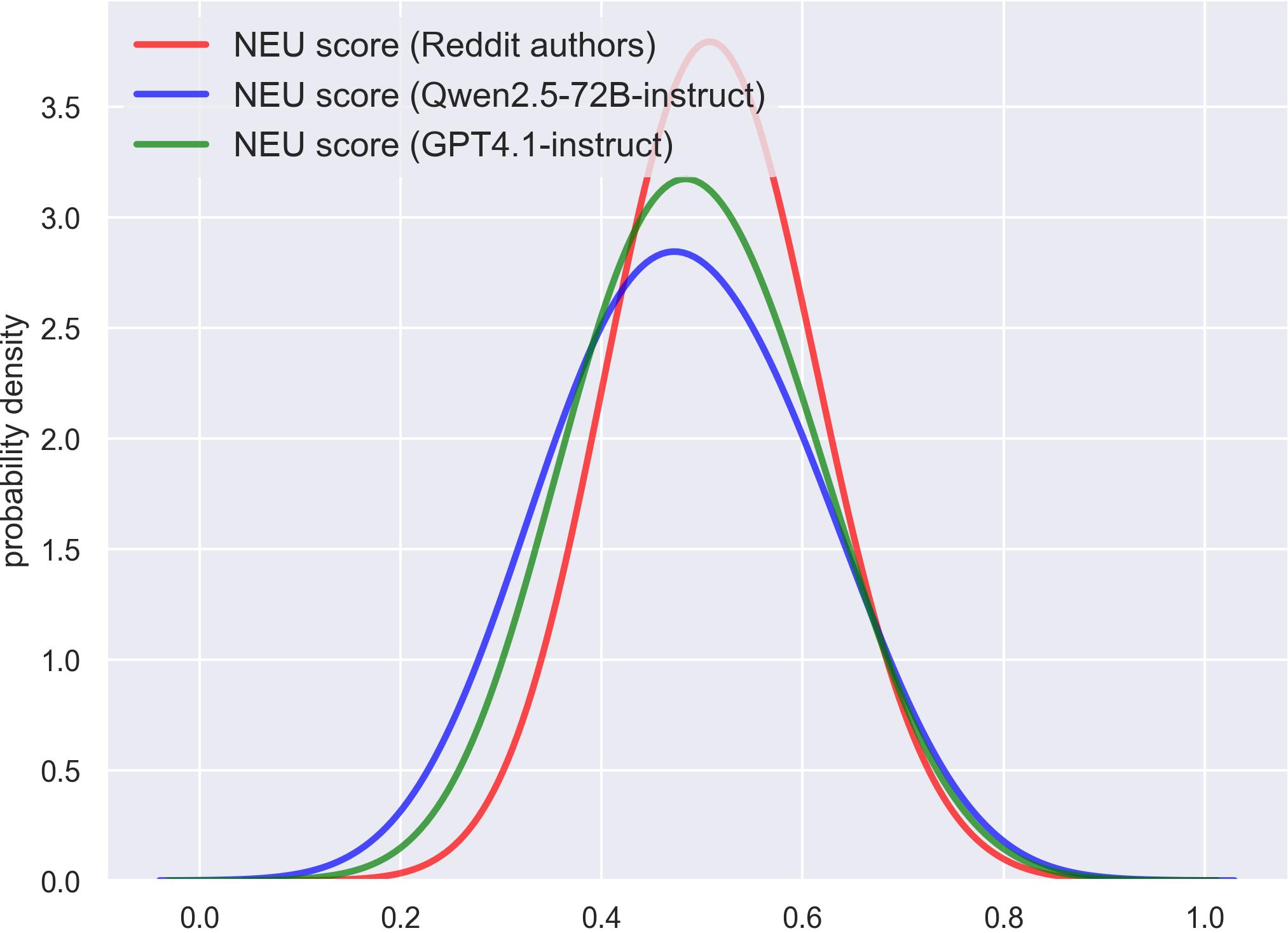}
\caption{Example kernel density distribution \\ of the NEU score in human comments vs LLMs: \\ differences are evident in both mean and STD.}
\label{fig:kernel-neu}
\end{minipage}
\end{figure}

Figures~\ref{fig:kernel-agr} and~\ref{fig:kernel-neu} further illustrate the kernel density distributions of the AGR and NEU traits in sample LLMs compared to human-authored comments. While Claude-Sonnet4.0 shows a distribution similar to that of Reddit authors, Llama3.3-70B exhibits a noticeably higher average, reflected as a right shift. For the NEU trait, the slight left shift of the two sample models reflects their relatively more "stable" nature compared to human writers.

Another notable observation in Table~\ref{tbl:results-personality} is that models show slightly higher STD values than human authors. This may be attributed to the broader range of personalities that models encounter in their training data, compared to the somewhat narrower fraction of the general population active on Reddit. We also observed no significant differences between the two temperature settings: results for $t{=}0.0$ are almost identical to those for $t{=}0.7$ across all models.
Finally, we assess the statistical significance of differences between humans and each model using two tests: the Mann-Whitney test for differences in the underlying distributions \citep{mann1947test}, and Levene's test for differences in variance \citep{levene1960robust}. Virtually all comparisons are significant at $p{<}0.01$; see Table~\ref{tbl:results-personality} for details.

\begin{table}[h!]
\caption{Big Five personality traits mean values ($\pm$ STD) for different models and human Reddit comments. We do report results for the CON trait here as well, but refrain from interpreting them. One careful observation would be that models show higher level of CON that human do in their comments. While EXT and OPN are generally comparable to those by human authors, considerably higher mean level of AGR and lower level of NEU is evident in text produced by LLMs. Virtually all models results show statistically significant different compared to humans --- for both underlying distributions (Mann-Whitney test), and variances (Levene test) --- at $p{<}0.01$. Results with no significant difference are marked with (-).}
\centering
\resizebox{1\textwidth}{!}{
\begin{tabular}{lrrrrrr}
\toprule
model name & \multicolumn{1}{c}{temp} & \multicolumn{1}{c}{EXT} & \multicolumn{1}{c}{OPN} & \multicolumn{1}{c}{AGR} & \multicolumn{1}{c}{NEU} & \multicolumn{1}{c}{CON} \\
\midrule
Claude-Sonnet4.0 & 0.0 & (-)0.490 ($\pm$0.086) & 0.440 ((-)$\pm$0.066) & 0.330 ($\pm$0.070) & 0.518 ($\pm$0.088) & 0.340 ($\pm$0.063) \\
Claude-Sonnet4.0 & 0.7 & (-)0.491 ($\pm$0.086) & 0.440 ($\pm$0.066) & 0.331 ($\pm$0.071) & 0.518 ($\pm$0.089) & 0.341 ($\pm$0.063) \\ \hdashline
GPT4.1 & 0.0 & 0.485 ($\pm$0.093) & 0.444 ($\pm$0.070) & 0.354 ($\pm$0.077) & 0.493 ($\pm$0.093) & 0.342 ($\pm$0.070) \\
GPT4.1 & 0.7 & 0.486 ((-)$\pm$0.092) & 0.443 ((-)$\pm$0.069) & 0.350 ((-)$\pm$0.075) & 0.498 ($\pm$0.092) & 0.342 ($\pm$0.068) \\ \hdashline
GPT4.1-mini & 0.0 & 0.481 ($\pm$0.093) & 0.453 ((-)$\pm$0.069) & 0.376 ($\pm$0.077) & 0.483 ($\pm$0.095) & 0.365 ($\pm$0.072) \\
GPT4.1-mini & 0.7 & 0.480 ((-)$\pm$0.095) & 0.454 ($\pm$0.069) & 0.377 ($\pm$0.079) & 0.484 ($\pm$0.095) & 0.366 ($\pm$0.073) \\
\midrule
Llama3.3-70B & 0.0 & (-)0.491 ($\pm$0.087) & 0.476 ($\pm$0.071) & 0.416 ($\pm$0.087) & 0.450 ($\pm$0.088) & 0.401 ($\pm$0.083) \\
Llama3.3-70B & 0.7 & (-)0.491 ($\pm$0.087) & 0.477 ($\pm$0.072) & 0.419 ($\pm$0.090) & 0.449 ($\pm$0.089) & 0.405 ($\pm$0.086) \\ \hdashline
Mixtral8x22B & 0.0 & 0.493 ($\pm$0.091) & 0.474 ($\pm$0.072) & 0.395 ($\pm$0.082) & 0.442 ($\pm$0.096) & 0.380 ($\pm$0.078) \\
Mixtral8x22B & 0.7 & 0.497 ($\pm$0.088) & 0.484 ($\pm$0.071) & 0.414 ($\pm$0.085) & 0.426 ($\pm$0.096) & 0.398 ($\pm$0.082) \\ \hdashline
Qwen2.5-72B & 0.0 & 0.486 ($\pm$0.096)& 0.472 ($\pm$0.074) & 0.400 ($\pm$0.081) & 0.476 ($\pm$0.102) & 0.376 ($\pm$0.076) \\
Qwen2.5-72B & 0.7 & 0.485 ($\pm$0.096) & 0.477 ($\pm$0.075) & 0.404 ($\pm$0.081) & 0.475 ($\pm$0.100) & 0.380 ($\pm$0.077) \\
\midrule
human authors & --- & 0.491 ($\pm$0.079) & 0.425 ($\pm$0.061) & 0.309 ($\pm$0.066) & 0.512 ($\pm$0.080) & 0.315 ($\pm$0.057) \\
\bottomrule
\end{tabular}
}
\label{tbl:results-personality}
\end{table}

\section{Generative AI: Gender}
\label{sec:experiments-gender}

Motivated by the findings of \citet{giorgi2021characterizing}, who observed that spambots exhibited very limited variation along demographic axes such as gender and age, we ask whether similar patterns can be observed in contemporary LLMs. Experimenting with multiple gender classifiers, we found that the DistilBERT-based classifier available on HuggingFace\footnote{\url{https://huggingface.co/padmajabfrl/Gender-Classification}} produced the most reasonable results, according to manual inspection. Each comment is assigned a continuous score in the 0–1 range, representing the likelihood that the comment was authored by a female, which we refer to as the "F-likelihood" score.

Table~\ref{tbl:user-text-gender-examples} presents example comments from our dataset (both human-written and generated), along with their automatically assigned F-likelihood scores. Results are shown for both open and closed models, and include posts containing comments with high (top) and low (bottom) F-likelihood scores.

\begin{table}[h!]
\caption{Example posts and their corresponding comments authored by human Redditors, as well as by closed and open LLMs. The probability score indicates the likelihood that the comment was written by a female author.}
\centering
\resizebox{\textwidth}{!}{
\begin{tabular}{lp{14cm}r}
\toprule
Reddit post & 100 girlfriends who really really really love you  is really really, really AWESOME. I started watching 100 girlfriends and man this anime is something i had just watched 1 episode when it aired and i though this seems boring, but [...] & \\ \midrule
Reddit author & Here's the thing: 100 Girlfriends, by its premise alone, should *not* work. At all. [...] Season 2 has somehow *surpassed* it in both hilarity and sincerity. I mean, the show is ridiculous, and it knows it most. So the fact that it can lean into it so hard with the constant fourth-wall breaks while still maintaining its genuine heart, I mean... it's incredible. & 0.999 \\ \midrule
Claude-Sonnet4.0 & Dude, you've barely scratched the surface! Episode 7 is just the beginning of the absolute chaos that is 100 Girlfriends. The Resident Evil parody had me dying too, especially when they committed so hard to the bit with the zombie makeup and everything. But wait until you get to some of the later episodes - there's a Dragon Ball reference that goes so hard I had to pause and rewind it three times. [...] This anime is criminally underrated and I'm so glad more people are discovering it. & 0.550 \\ \midrule
Qwen2.5-72B & I can't believe I'm saying this, but you've got me intrigued! I initially thought 100 Girlfriends would be just another harem anime, but it seems like it’s got a lot more depth and humor than I expected. The references to other media, like God of War, really add a fun layer to the show. I’ve been a Kratos fan for a long time, and the idea of [...] & 0.673 \\ \midrule \midrule
Reddit post & What’s going on with DeepSeek? Seeing things like [...] in regards to DeepSeek. Isn’t it just another LLM? I’ve seen other posts around how it could lead to the downfall of Nvidia and the Mag7? Is this just all bs? & \\ \midrule
Reddit author & DeepSeek, a Chinese AI startup, just dropped its R1 model, and it’s giving Silicon Valley a panic attack. Why? They trained it for just \$5.6 million, chump change compared to the Billions companies like OpenAI and Google throw around, and are asking the US government for Billions more. The silicon valley AI companies have been saying that there's no way to train AI cheaper, and that what they need is more power. [...] & 0.001 \\ \midrule
GPT4.1 & I think a lot of the hype around DeepSeek is classic social media FOMO and misunderstanding of how these AI models actually work. DeepSeek is a large language model, yes, but it’s not fundamentally different from what OpenAI, Google, or Meta have been developing. The claims that it could lead to the "downfall" of Nvidia or the Magnificent 7 (Apple, Microsoft, Alphabet, Amazon, Meta, Tesla, Nvidia) seem pretty far-fetched. Most of these companies have deeply entrenched infrastructure [...] I'd take all these doom-and-gloom or utopian predictions with a huge grain of salt. & 0.107 \\ \midrule
Mixtral8x22B & DeepSeek is indeed based on a large language model (LLM), but it's not just another LLM. It's designed to go beyond mere language generation and has the potential to revolutionize various industries, including AI and hardware manufacturing. The concerns about Nvidia and the Mag7 might be overblown, but it's worth noting that such innovations could disrupt the status quo. However, it's still in its early stages, and only time will tell its true impact. [...] & 0.001 \\
\bottomrule
\end{tabular}
}
\label{tbl:user-text-gender-examples}
\end{table}

\begin{figure}[h]
\centering
\includegraphics[width=\linewidth]{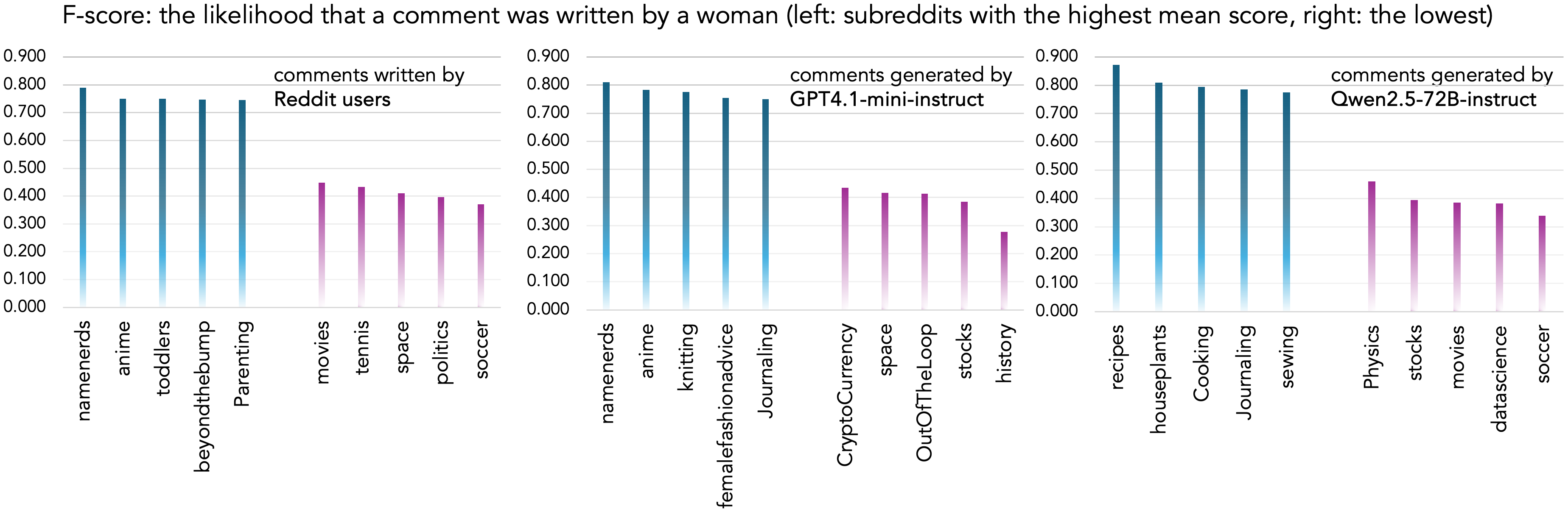}
\caption{Subreddits exhibiting the highest and the lowest mean F-likelihood score in comments produced by Reddit human authors and models.}
\label{fig:gender-highest-lowest}
\end{figure}

\begin{table}[h!]
\caption{The mean likelihood that the comment was written by a female (F) author, the ratio of F-authored comments (those with F-likelihood exceeding 0.5), the ratio of M-authored comments. The standard deviation is specified in parenthesis. Generative models are compared to results produced by human users. Virtually all models results show statistically significant difference compared to humans --- for both underlying distributions (Mann-Whitney test), and variances (Levene test) --- at $p{<}0.01$. Results with no significant difference are marked with (-).}
\centering
\begin{tabular}{lrrrr}
\toprule
model & \multicolumn{1}{c}{temp} & mean F-likelihood & comments' F-ratio & comments' M-ratio \\
\midrule
Claude-Sonnet4.0 & 0.0 & 0.535 ($\pm$0.282) & 0.567 & 0.433 \\
Claude-Sonnet4.0 & 0.7 & 0.532 ($\pm$0.282) & 0.558 & 0.442 \\ \hdashline
GPT4.1 & 0.0 & (-)0.602 ($\pm$0.256) & 0.700 & 0.300 \\
GPT4.1 & 0.7 & (-)0.589 ($\pm$0.260) & 0.678 & 0.322 \\ \hdashline
GPT4.1-mini & 0.0 & 0.580 ($\pm$0.260) & 0.667 & 0.333 \\
GPT4.1-mini & 0.7 & 0.582 ($\pm$0.257) & 0.666 & 0.334 \\
\midrule
LLama3.3-70B & 0.0 & 0.553 ($\pm$0.242) & 0.621 & 0.379 \\
LLama3.3-70B & 0.7 & 0.548 ($\pm$0.242) & 0.612 & 0.388 \\ \hdashline
Mixtral8x22B & 0.0 & 0.639 ($\pm$0.251) & 0.741 & 0.259 \\
Mixtral8x22B & 0.7 & 0.628 ($\pm$0.251) & 0.728 & 0.272 \\ \hdashline
Qwen2.5-72B & 0.0 & (-)0.600 ($\pm$0.264) & 0.700 & 0.300 \\
Qwen2.5-72B & 0.7 & (-)0.612 ($\pm$0.262) & 0.712 & 0.287 \\
\midrule
human authors & --- & 0.591 ($\pm$0.300) & 0.656 & 0.344 \\
\bottomrule
\end{tabular}
\label{tbl:results-gender}
\end{table}

\subsection{Evaluation of Gender Classification Results}
We further validate the automatically assigned F-likelihood scores by identifying five subreddits with the highest and lowest mean scores. This computation was performed for both human- and model-generated comments. Figure~\ref{fig:gender-highest-lowest} illustrates the results: careful inspection shows that the findings largely align with intuition. Among Redditors, comments likely written by female authors are concentrated in threads such as \texttt{namenerds}, \texttt{toddlers}, \texttt{beyondthebump} (motherhood), \texttt{anime}, and \texttt{Parenting}. Similarly, LLM-generated comments display plausible gender patterns, with \texttt{knitting}, \texttt{femalefashionadvise}, \texttt{sewing}, and \texttt{Cooking} appearing among the subreddits with high F-likelihood. Subreddits with low F-likelihood scores (i.e., high M-likelihood) are consistently associated with topics like \texttt{politics}, \texttt{soccer}, \texttt{stocks}, and \texttt{movies}. We conclude that F-likelihood score assignments are sufficiently reliable, and perform human- vs models comparative analysis.

\subsection{Gender: Human Authors vs Generative Models}
We compute the mean comment F-likelihood score for texts written by humans and those generated by LLMs. Table~\ref{tbl:results-gender} reports the mean and standard deviation (STD) results. The models exhibit a range of mean scores around the average F-likelihood of 0.591 observed in human comments: some LLMs show slightly lower averages, while others are slightly higher, with no consistent pattern. A systematic difference is evident in the STD values: models display lower variance, indicating slightly more limited variation in gendered language, consistent with the findings on spambots by \citet{giorgi2021characterizing}.

As before, we assess the significance of differences between humans and each model using two statistical tests: the Mann-Whitney test for differences in the underlying distributions, and Levene's test for differences in variance. All differences except those of GPT4.1 and Qwen2.5-72B for underlying distributions are significant at $p{<}0.01$; see Table~\ref{tbl:results-gender} for further details.

\section{Conclusions}
\label{sec:conclusions}

In this study, we examined the personality and gender characteristics of texts produced by contemporary LLMs in comparison to human-written comments on Reddit. Using established personality and gender classifiers, we analyzed thousands of posts and comments, observing both similarities and systematic differences. Our results indicate that models can capture many human-like patterns for traits such as Extroversion and Openness, while systematically exhibiting higher Agreeableness and lower Neuroticism, reflecting their cooperative and psychologically stable training objectives. Similarly, gendered language in model-generated text broadly aligns with human patterns, though models show slightly reduced variation, echoing previous observations in social spambots.

Overall, these findings suggest that current LLMs are capable of producing text that mirrors some aspects of human personality and demographics, while also highlighting consistent divergences that reflect model design and training biases. 

\section*{Ethical Considerations}
\label{sec:ethical}

Here we address the main concern of anonymity of Reddit users. Data used for this research can only be associated with participants' user IDs, which, in turn, cannot be linked to any identifiable information, or used to infer any personal or demographic trait. \citet{jagfeld2021understanding} debated the need to obtain informed consent for using social media data mainly because it is not straightforward to determine if posts pertain to a public or private context. Ethical guidelines for social media research \citep{benton2017ethical} and practice in comparable research projects \citep{ahmed2017using}, as well as {Reddit's terms of use},\footnote{\url{https://www.redditinc.com/policies/user-agreement-september-12-2021}} regard it as acceptable to waive explicit consent if users' anonymity is protected.

We reinforce that our dataset does not contain user IDs for neither posts nor comments. This data can be retrieved using a post or comments ID, which is attached to each text in the dataset.

\section*{Limitations}
\label{sec:limitations}

We acknowledge several limitations of our study. First, our analyses rely on automatic classifiers for personality and gender as if these constituted ground truth. These classifiers, however, are themselves trained on human-generated and limited data --- Reddit discussions, in our case --- and inevitably reflect the social, cultural, and methodological biases embedded in their training sources. A related concern is our use of Reddit as the human baseline: although it offers a large and accessible corpus, it is not necessarily representative of the broader population and carries its own community-specific norms and cultural biases. Furthermore, our work is restricted to English, which limits the generalizability of the findings to other languages and cultural contexts.

Another limitation concerns the interpretation of what the classifiers' scores reveal about LLMs. It remains unclear whether these scores capture any intrinsic properties of the models or merely reflect surface-level stylistic regularities. Drawing a clearer conceptual boundary between stylistic tendencies and psychological constructs would be our focus in future work.
Finally, our prompts to the models, while aiming at simulating a Reddit user, may cause undesired biases: instructing models to "behave like several social media users", for instance, may itself bias stylistic patterns in ways that confound the personality and gender inferences drawn by the classifiers. Future work should aim to disentangle such prompt-induced effects from the model's inherent linguistic tendencies.

\begin{acknowledgments}
We are grateful to our four anonymous reviewers for their useful comments and constructive feedback.
\end{acknowledgments}

\section*{Declaration on Generative AI}
The authors have not employed any Generative AI tools while writing the paper.

\bibliography{custom}

\end{document}